\documentclass[runningheads]{llncs}

\usepackage{cite}
\usepackage{graphicx}
\usepackage{subcaption}
\usepackage{amsmath,amssymb}
\usepackage{multirow}
\usepackage{hyperref}
\usepackage{pdfpages}

\usepackage[utf8]{inputenc}

\begin{document}
\title{Unsupervised Anomaly Localization using Variational Auto-Encoders}
%
%
\author{David Zimmerer\inst{1},
Fabian Isensee\inst{1},
Jens Petersen\inst{1},
Simon Kohl\inst{1},\\
Klaus Maier-Hein\inst{1}}
%
\authorrunning{D. Zimmerer et al.}
%
\institute{German Cancer Research Center (DKFZ), Heidelberg, Germany
}
\maketitle              
\begin{abstract}
An assumption-free automatic check of medical images for potentially overseen anomalies would be a valuable assistance for a radiologist. Deep learning and especially Variational Auto-Encoders (VAEs) have shown great potential in the unsupervised learning of data distributions. In principle, this allows for such a check and even the localization of parts in the image that are most suspicious. Currently, however, the reconstruction-based localization by design requires adjusting the model architecture to the specific problem looked at during evaluation. This contradicts the principle of building assumption-free models. We propose complementing the localization part with a term derived from the Kullback-Leibler (KL)-divergence. For validation, we perform a series of experiments on FashionMNIST as well as on a medical task including $>$1000 healthy and $>$250 brain tumor patients. Results show that the proposed formalism outperforms the state of the art VAE-based localization of anomalies across many hyperparameter settings and also shows a competitive max performance.



\end{abstract}
\section{Introduction}


Unsupervised anomaly detection is a key technique that could allow us to overcome the data bottleneck that is ever so present especially in the medical domain. Unsupervised models can directly learn the data distribution from a large cohort of unannotated subjects and then be used to detect out of distribution samples and thus ultimately identify diseased or suspicious cases. By decoupling abnormality detection from reference annotations, these approaches are completely independent of human input and can therefore be applied to any medical condition or image modality. 
First approaches on unsupervised anomaly detection were built on explicit assumptions, often preventing generalizability to other tasks. Juan-Abarrachin et al. \cite{juan-albarracin_automated_2015} manually designed a set of image features for brain tumor detection. By mapping the original image into carefully chosen feature representations, they were able to separate tumorous from healthy tissue by clustering of the voxels in feature space and combining this with an atlas-based approach. Erihov et al. \cite{erihov_cross_2015} instead relied on the natural symmetry of the brain (as well as some other organs) to identify regions that behave abnormally.
Only the era of deep-learning has allowed to address the problem in a more principled fashion, with the aim of learning normal data distributions in order to detect abnormal samples. Ideally, anomaly detection should not build upon case-specific assumptions in the form of medical domain knowledge or specific annotated validation sets to optimize for, which should be interpreted as an unwanted form of supervision implicitly added by design of the method. 
Variational Auto-Encoders (VAEs) and their extensions are, alongside flow-based and auto-regressive models, a current de-facto standard for density estimation and particularly anomaly/out-of-distribution sample detection tasks \cite{abati_and:_2018,an_variational_2015,kiran_overview_2018,nalisnick_deep_2018}. Here, the evidence lower bound (ELBO), by definition a combination of the reconstruction error with the Kullback-Leibler (KL)-divergence, commonly serves as a proxy for the sample likelihood \cite{an_variational_2015,kiran_overview_2018}. 
Recent work has even demonstrated the ability of VAEs to localize and segment the parts in the image that are most suspicious, which is of particular importance in medical applications. Pawlowski et al. \cite{pawlowski_unsupervised_2018} compare different auto-encoders for CT based pixel-wise segmentation and Baur et al. \cite{baur_deep_2018} propose a VAE with an adversarial loss on the reconstructions to improve performance.
Chen et al. \cite{chen_deep_2018,chen_unsupervised_2018} use a VAE with an additional adversarial latent loss. 
The localization part in these studies is currently solely based on the reconstruction error, thus outlining regions as suspicious if they cannot be adequately reconstructed by the model. 
You et al. \cite{you_unsupervised_2019} include the KL-term for reconstructions closer to the data manifold.
In this work, we demonstrate that reconstruction-based anomaly detection is sub-optimal. One obvious deficiency is that the capability of a VAE to reconstruct anomalies is by design tightly coupled to the expressiveness (size and configuration) of the latent space. This, at the same time, also explains why reconstruction-based techniques are still able to achieve high performance scores on unsupervised tasks: their deficiencies can be compensated for to a certain extent by tuning the model architecture towards being optimally suited for a specific tasks (see also \cite{chen_deep_2018,pawlowski_unsupervised_2018,goldstein_comparative_2016}), as is common practice when hyperparameters are optimized on annotated validation sets. 
Task specific hyperparameter optimization, however, contradicts the principle of assumption-free anomaly detection. 
To investigate this we analyze the robustness of the reconstruction-based anomaly detection on a sample-wise and pixel-wise level and compare it to the ELBO and the KL-divergence.
Inspired by the results, we propose to integrate the KL-divergence of a VAE into a pixel-wise anomaly detection as well. This is in analogy to sample-wise anomaly detection, where the ELBO is also based on both the reconstruction error and the KL-divergence.
The proposed approach outperforms reconstruction-based localization in a broad variety of different model configurations, demonstrating the robustness with respect to hyperparameters. 
When allowing task-specific fine-tuning, the model as well outperforms previously reported deep-learning based results.

\section{Methods}

\subsection{Variational Autoencoders for anomaly detection}

VAEs can approximate data distributions by optimizing a lower bound   $\mathcal{L}$, often termed evidence lower bound (ELBO) \cite{kingma_auto-encoding_2013,rezende_stochastic_2014,dai_diagnosing_2019}. It is defined as 

\begin{equation}
    \label{eq:elbo}
    \log p(x) \geq \mathcal{L} = -D_{KL}(q(z|x) || p(z)) + \mathbb{E}_{q(z|x)} [\log p(x|z)],
\end{equation}

where $q(z|x)$ and $p(x | z)$ are diagonal normal distributions parameterized by neural networks $f_{\mu}$, $f_{\sigma}$, and $g_{\mu}$ and constant $c$:

\begin{equation}
    \begin{split}
        & q(z|x) = \mathcal{N}(z; f_{\mu, \theta_1}(x), f_{\sigma, \theta_2}(x)^2 ),\\
        & p(x | z)=\mathcal{N}(x ; g_{\mu, \gamma}(z), \mathbb{I}*c), 
    \end{split}
\end{equation}

However VAEs have design choices and data dependent parameters which can influence the performance, such as the network architecture, the number of latent variables, the standard deviation $c$ of $p(x | z)$ and the data dimension. 
Given sufficiently powerful neural networks $f_{\mu}$, $f_{\sigma}$, and $g_{\mu}$ and a large enough latent space, VAEs with Gaussian encoders and decoders can (and under some conditions will \cite{dai_diagnosing_2019}) approximate the true data distribution.
Thus, after optimization, $\mathcal{L}$ is often used as a proxy for the likelihood of a data sample and consequently as an anomaly score.

\subsection{Pixel-wise anomaly detection}
\label{ssec:panno}

For medical applications a pixel-wise localization, similar to a segmentation, is often more desirable than a sample-wise score. 
Related methods typically generate a segmentation map by thresholding the pixel-wise reconstruction error \cite{baur_deep_2018,chen_unsupervised_2018,chen_deep_2018,pawlowski_unsupervised_2018,schlegl_unsupervised_2017}.
However, in contrast to anomaly detection in the common sample-wise setting, this discards the KL-term and potentially ignores useful information, since a low $\mathcal{L}$ and consequently high anomaly score can be caused by the reconstruction-term ($\mathbb{E}_{q(z|x)} [\log p(x|z)]$) and/or the KL-term ($D_{KL}(q(z|x) || p(z))$).
To alleviate this problem we aim to include, in addition to the pixel-wise reconstruction error (``\textbf{Rec-Error}''), the KL-term for a pixel-wise anomaly scoring. Our experiments include different strategies of achieving this as well as strategies for each term separately:
\begin{itemize}
    \item ``\textbf{ELBO-grad}'': Building on the assumption that $\mathcal{L}$ allows for a good enough approximation of the true data distribution, we propose to use the derivative of $\mathcal{L}$ with respect to the input, yielding a pixel-wise vector pointing towards a data sample with a lower $\mathcal{L}$:
    
    \begin{equation}
        \label{eq:score}
        \textit{ELBO-Grad}_{score}(x_i) = |[\frac{\partial \mathcal{L}}{\partial x}]_i|.
    \end{equation}
    
    Given that $\mathcal{L}$ is locally convex, the magnitude of the pixel gradient should correspond to a pixel-wise anomaly score \cite{alain_what_2014}.
    \item  ``\textbf{KL-grad}'': To get a pixel-wise score for only the KL-term of $\mathcal{L}$, we differentiate the KL-term with respect to the input.
    \item ``\textbf{Rec-Grad}'': To get a pixel-wise score for only the reconstruction-term of $\mathcal{L}$, we differentiate the reconstruction-term of $\mathcal{L}$ with respect to the input.
    \item ``\textbf{Combi}'': Instead of differentiating the reconstruction-term of $\mathcal{L}$, we can directly use the reconstruction error and combine it with ``\textit{KL-grad}''. This should be less prone to noise artifacts. 
    For this model, we combine the derivative of the KL-term with the reconstruction error by multiplication, since the terms differ by several orders of magnitude.
\end{itemize}


\subsection{ Generalization and Robustness across different parameters }

We compare the discriminative performance of the ELBO $\mathcal{L}$, the KL-term, and the reconstruction-term separately to inspect the information they contain about the data distribution and consequentially the abnormality.
We further analyze the robustness and generalizability across different parameter settings and different (medical and non-medical) datasets.

First, we use the FashionMNIST dataset \cite{xiao_fashion-mnist:_2017}, where we train and validate the model on 54000 images using 9 out of the 10 provided classes and then evaluate the performance by attempting to discriminate between the classes seen during training and the 10th unseen class. In analogy to \cite{paszke_automatic_2017}, we used a model with a 3-layer fully connected encoder and decoder with 400 hidden units and ReLU non-linearities. 
To analyze the robustness we vary the number of latent variables, the standard deviation $c$ of $p(x|z)$ (resulting in a down or up-weighing of the reconstruction loss), the image-size/scaling and class left out during training.
By default we use 20 latent variables, $c=1$, a scaling factor of 1 and class $0$ is left out during training.

Second, we train and validate on the patient data from the HCP dataset \cite{van_essen_human_2012} ($N=1000$ patients) and test on the BraTS2017 dataset \cite{menze_multimodal_2015} ($N=250$ patients). While the HCP patients are all young healthy subjects, the patients in the BraTS dataset all have brain tumours. Finding tumours in this setting is a particularly hard problem because additionally to the challenging nature of the task itself there is a considerable domain shift between the datasets.
The BraTS dataset includes tumor annotations which we can use for model evaluation. We treat slices without annotations as healthy whereas slices with at least 20 annotated tumor voxels are considered diseased. Our model consists of a 5-Layer fully-convolutional encoder and decoder with feature-map size of 16-32-64-256. We use strided convolutions (stride 2) as downsampling and transposend convolutions as upsampling operations, each followed by a LeakyReLU non-linearity (inspired by DCGAN \cite{radford_unsupervised_nodate}).
To analyze the robustness and generalizability of the different methods, we vary the number of latent variables (default 256), the standard deviation $c$ of $p(x|z)$ (default 1) and the image-size (default $64\times64$ pixels).

The models are trained with Adam and an initial learning rate of $0.0001$. 
Whenever the validation loss $\mathcal{L}_{val}$ reaches a plateau, the learning rate is decreased by multiplying it with $0.1$. 
The training is stopped once the validation loss does not decrease for more than 3 epochs.
For each model we perform 5 runs and report the mean as well as the max/min performance. 
The code to replicate the results is available at \href{https://github.com/MIC-DKFZ/vae-anomaly-experiments}{https://github.com/MIC-DKFZ/vae-anomaly-experiments}.

\section{Results}

\subsection{Sample-wise performance}
\label{ssec:robust}

\begin{figure}[tb]
 \centering
 \begin{subfigure}[t]{0.22\textwidth}
    \includegraphics[width=0.99\textwidth]{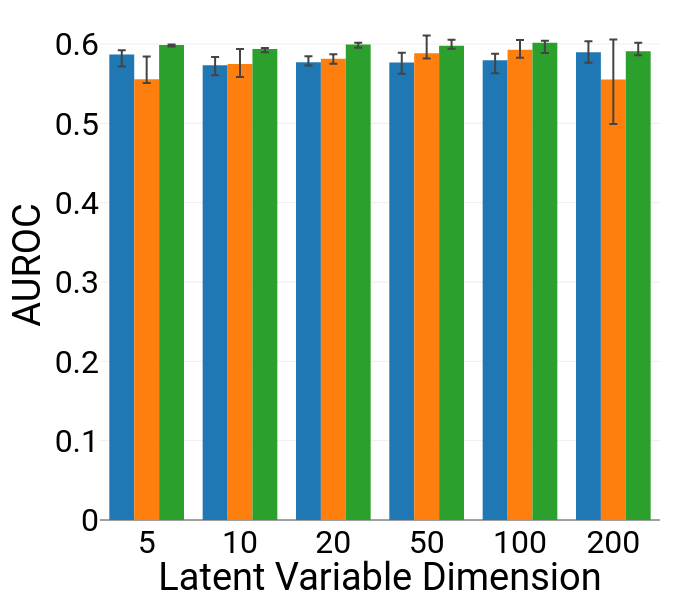}
 \end{subfigure}
 \begin{subfigure}[t]{0.22\textwidth}
    \includegraphics[width=0.99\textwidth]{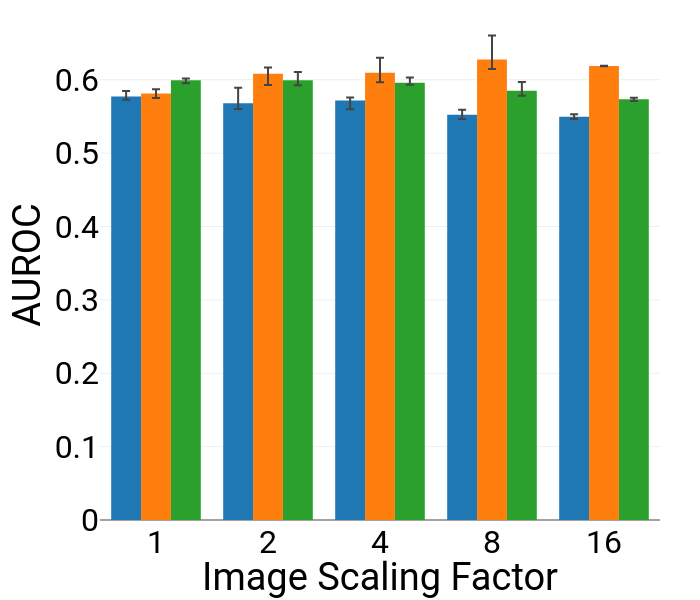}
 \end{subfigure}
 \begin{subfigure}[t]{0.22\textwidth}
    \includegraphics[width=0.99\textwidth]{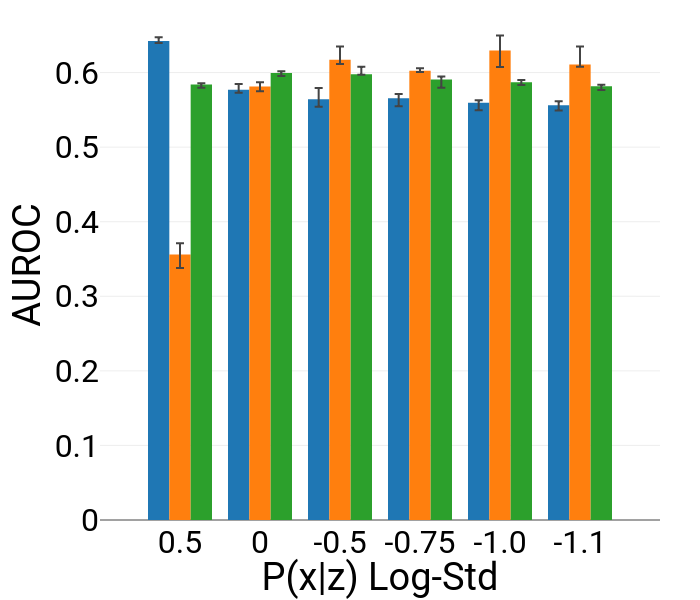}
 \end{subfigure}
 \begin{subfigure}[t]{0.26\textwidth}
    \includegraphics[width=0.99\textwidth]{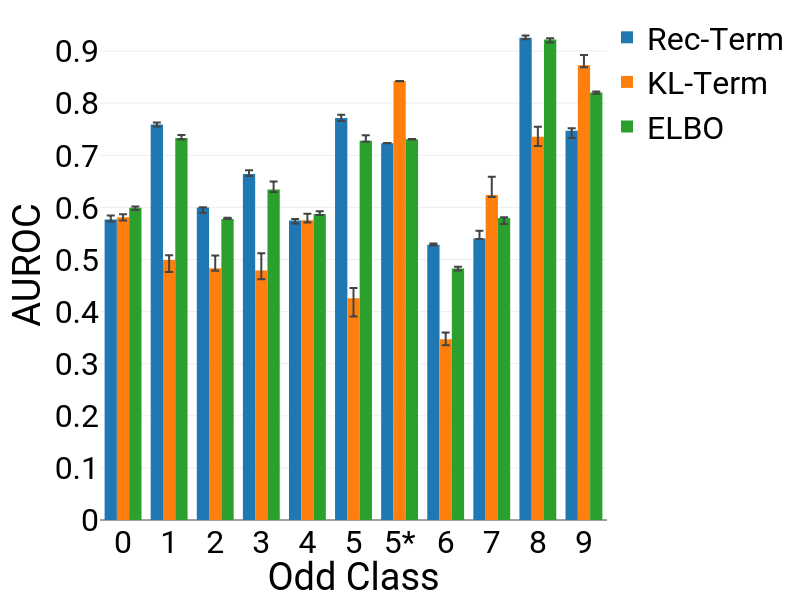}
 \end{subfigure}
 \\
 \begin{subfigure}[t]{0.22\textwidth}
   \includegraphics[width=0.99\textwidth]{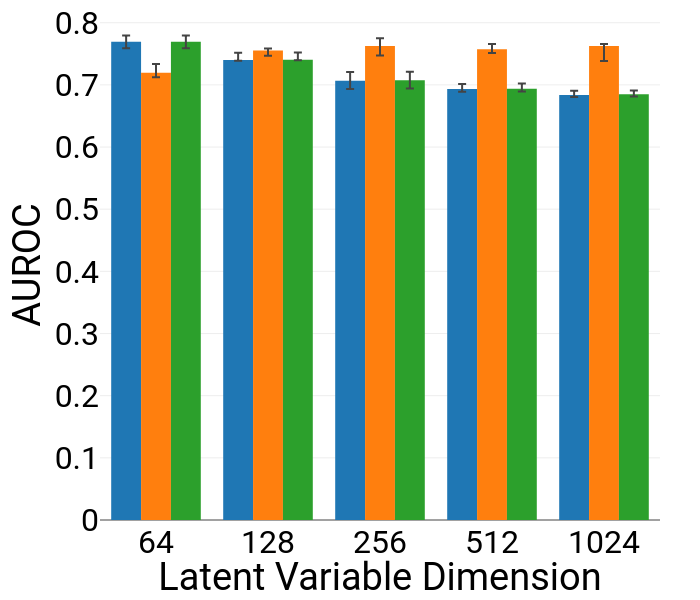}
 \end{subfigure}
 \begin{subfigure}[t]{0.22\textwidth}
   \includegraphics[width=0.99\textwidth]{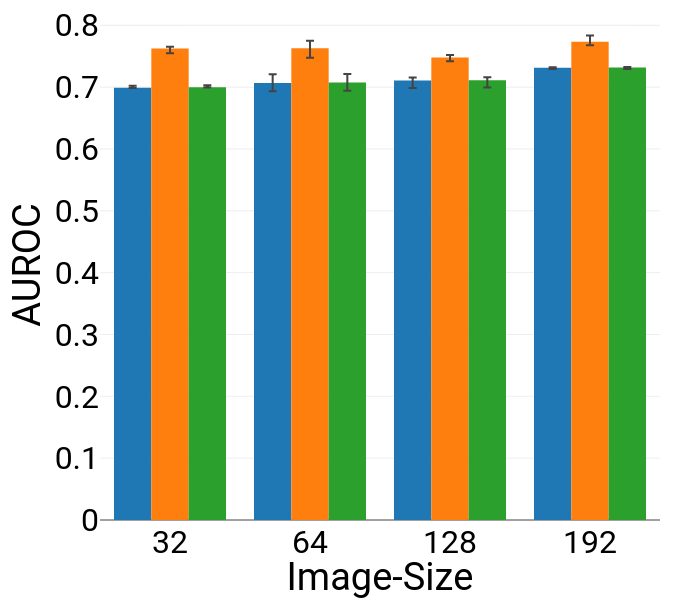}
 \end{subfigure}
 \begin{subfigure}[t]{0.26\textwidth}
   \includegraphics[width=0.99\textwidth]{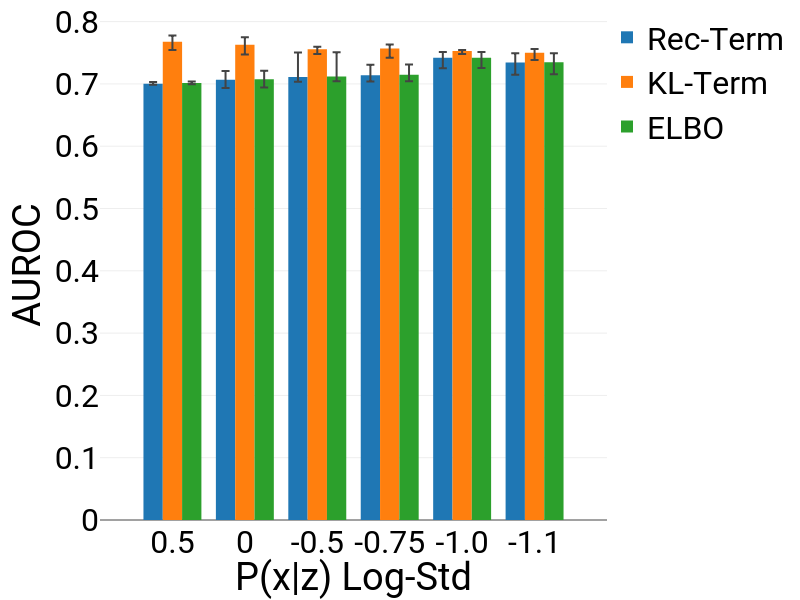}
 \end{subfigure}
 \caption{Sample-wise anomaly detection AUROC for reconstruction-term (Rec), the KL-term, and the ELBO $\mathcal{L}$ for the FashionMNIST dataset (first row) and the BraTS2017 dataset (second row) over different VAE design choices. $5^\star $ shows a fine-tuned performance with odd-class 5 ($\log c = 1.4$). }
 \label{fig:sample}
\end{figure}

The sample-wise results across different parameter settings can be seen in Fig. \ref{fig:sample}. 
It is apparent that in most cases the reconstruction-term shows lower discriminative power than either the KL-term or the ELBO. Consequently, important information is lost when focusing only on the reconstruction error. Furthermore, cases where it has better performance, the model is severely constrained, for example by having a small latent variable dimension (which was shown in \cite{dai_diagnosing_2019} to hinder VAEs from approximating the data distribution and to lead to poor reconstruction). Thus the robustness of the KL-term can perhaps be intuitively explained by \cite{dai_diagnosing_2019}, which states that for VAEs the ELBO best approximates the data distribution having ``perfect reconstructions using the fewest number of clean, low-noise latent dimensions''.
So far, no hyperparameters were specifically tuned to specifically improve one of the losses.
However we want to demonstrate that by using an annotated validation set to tune the parameters, the performance of the reconstruction error as well as the KL-term individually can give competitive performance. 
This is done by choosing the odd class with the largest gap between the rec-loss and the kl-loss (class 5) and using a single hyperparameter adjustment (setting $\log c = 1.4$). By doing so, we were able to achieve an area under the receiver operator curve (AUROC) of $0.82$ for the KL-loss, which now significantly outperformed the reconstruction loss.
However, when no annotated dataset is available our results indicate that in general including the KL-loss for anomaly scoring not just for a sample-wise level, but also on a pixel-wise level could increase performance, as analyze next.

\subsection{ Pixel-wise performance }

\begin{figure}[tb]
 \centering
 \begin{subfigure}[t]{0.22\textwidth}
    \includegraphics[width=0.99\textwidth]{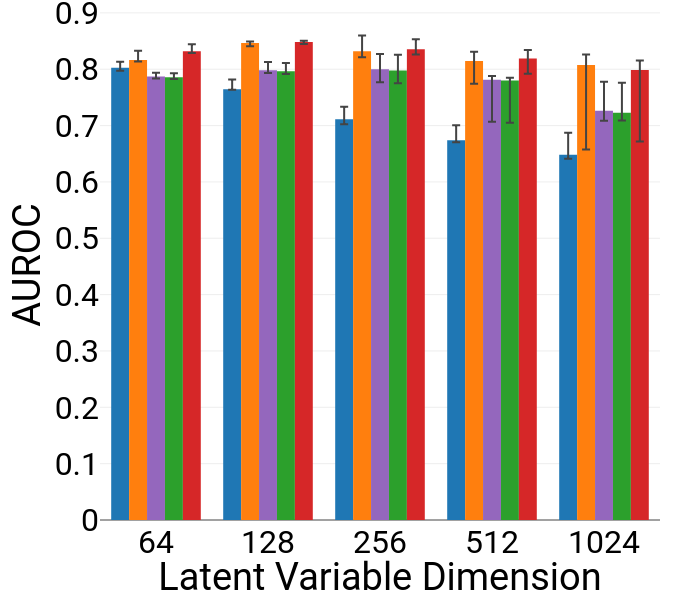}
 \end{subfigure}
 \begin{subfigure}[t]{0.22\textwidth}
    \includegraphics[width=0.99\textwidth]{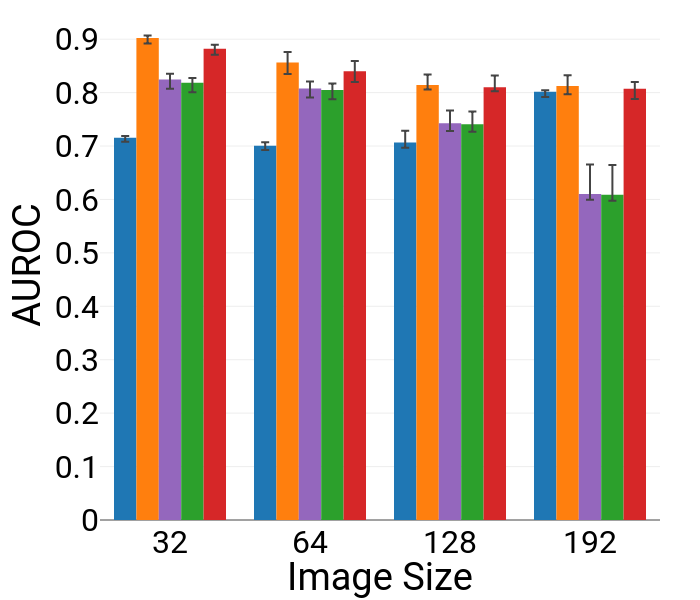}
 \end{subfigure}
 \begin{subfigure}[t]{0.26\textwidth}
    \includegraphics[width=0.99\textwidth]{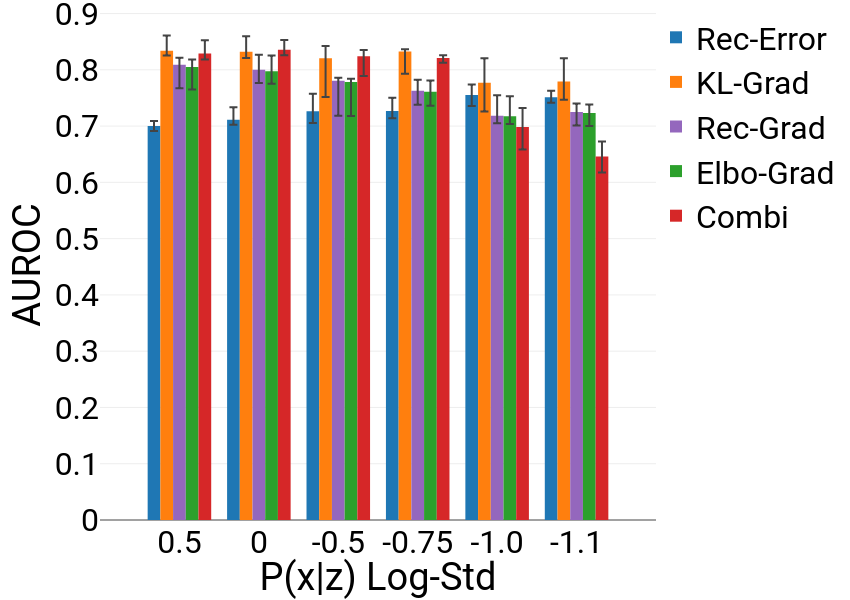}
 \end{subfigure}
 \caption{Pixel-wise AUROC over different VAE design choices on the BraTS2017 dataset. We compare the reconstruction loss, the KL-term gradient, the reconstruction-term gradient, the ELBO $\mathcal{L}$ gradient and our \textit{combi} method.}
  \label{fig:pixel}
\end{figure}

The pixel-wise performance on the BraTS2017 dataset across different hyperparamater settings is summarized in Fig. \ref{fig:pixel}. Here the model was trained on the healthy HCP subjects and then applied to BraTS2017 for anomaly detection.
We used the same model and data setting as before, but in this case evaluate the performance to detect pixel-wise whole tumor annotations.
We compare the methods presented in Sec. \ref{ssec:panno}: The pixel-wise reconstruction-error (\textit{Rec-Error}), the backpropagated $\mathcal{L}$ (\textit{Elbo-Grad}), its backpropagated KL-term (\textit{KL-Grad}) and reconstruction-term (\textit{Rec-Grad}) separately as well as the \textit{combi} model. 
Similar to the previous cases, it is obvious that in most cases the reconstruction error alone is outperformed by other methods. 
Furthermore, for most choices the \textit{KL-Grad} and the \textit{combi} model perform best, indicating a more robust performance at least for this particular dataset (similar observations can be made for the ISLES2015 dataset, as shown in the Suppl.).



\subsection{ Hyperparameter tuning }

\begin{figure}[tb]
\centering
\begin{subfigure}{0.44\textwidth}
\includegraphics[width=0.99\textwidth]{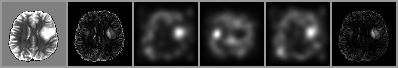}
\end{subfigure}
\begin{subfigure}{0.44\textwidth}
\includegraphics[width=0.99\textwidth]{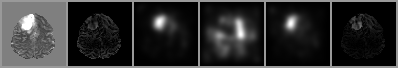}
\end{subfigure}
\begin{subfigure}{0.44\textwidth}
\includegraphics[width=0.99\textwidth]{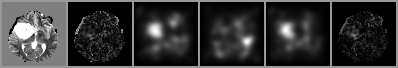}
\end{subfigure}
\begin{subfigure}{0.44\textwidth}
\includegraphics[width=0.99\textwidth]{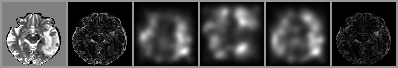}
\end{subfigure}
\begin{subfigure}{0.44\textwidth}
\includegraphics[width=0.99\textwidth]{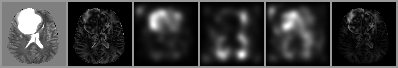}
\end{subfigure}
\begin{subfigure}{0.44\textwidth}
\includegraphics[width=0.99\textwidth]{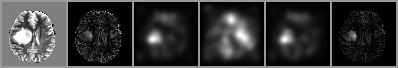}
\end{subfigure}
\caption{Anomaly detection of the fine-tuned model on six test set samples. For each example, the reconstruction-error, the backpropagated KL-term, the backpropagated reconstruction-term, the backpropagated $\mathcal{L}$ and the \textit{combi} is plotted from left to right.}
\label{fig:images}
\end{figure}

The top performing methods in the experiments shown above already exhibit high AUROC $>0.9$. 
In particular, the often ignored KL-term shows robust performance across a variety of tested settings. The \textit{combi} approach exhibits a similar robustness.
We were interested in investigating the top performance of the KL-term approach in a scenario where an annotated validation set can be employed for hyperparameter tuning, just as it is often done in the literature when presenting reconstruction-based approaches.
Results are shown in Table \ref{tab:dice} and Fig. \ref{fig:images}.
Dice scores are calculated by thresholding the anomaly values at a value that was determined using $\frac{1}{5}$ of the test dataset. The reported dice scores were then taken from the other $\frac{4}{5}$th of the dataset.


\begin{table}[tb] 
\centering
\begin{tabular}{|c|c|c|c|c|c|c|}
\hline
                                \multicolumn{2}{|c|}{deep-learning} &
                            \multicolumn{2}{|c|}{ours} &
                            
                            \multicolumn{3}{|c|}{non deep-learning} 
                        \\
\hline
                         $\alpha$-GAN \cite{chen_deep_2018}& VAE-Rec \cite{chen_deep_2018}& default &  fine-tuned &
                            GHMRF \cite{juan-albarracin_automated_2015} & X-Saliency \cite{erihov_cross_2015} & GMM \cite{chen_deep_2018}  \\
                           (15) &  (15) & (17) & (17) & (13) & (HGG 14)& (15)\\ \hline
                              0.37         &    0.42      &     0.36 &     0.44 & 0.72    &  0.75      &   0.22               \\ \hline
\end{tabular}
\caption{Dice of unsupervised whole tumor detection on the BraTS Dataset (the number in brackets specifies the year of the BraTS dataset that was used). While our approach is outperformed by non-deep learning approaches that were specifically designed with domain knowledge of the dataset in mind, it is competitive with other deep-learning based anomaly detection approaches. }
\label{tab:dice}
\end{table}

\section{Discussion \& Conclusion}

In this work we compared different approaches of detecting anomalies with VAEs over many different hyperparameter settings. However, a hyperparameters are regularly chosen by task specific optimization on an annotated validation set, which contradicts the principle of unsupervised anomaly detection. We showed that for a pixel-wise anomaly detection the reconstruction error does not always have the best performance and can regularly be improved by combining it with the backpropagated KL-term. This is in analogy to common VAE-based anomaly detection methods that consider the ELBO $\mathcal{L}$, which constitutes definition the combination of the KL-term with the reconstruction-term, as a proxy score. The proposed approaches shows promising performance across a broad range of hyperparameters and thus effectively reduce the need for manually tuning towards a validation set and thus keep the unsupervised property intact. If an annotated validation set is available, however, our approach can still be fine tuned to achieve the same competitive performance as other methods. 
On a first glance, our method is outperformed by non-deep learning methods on the BraTS dataset. This however neglects the fact that these models incorporate specific domain knowledge via their algorithmic design. While this results in a strong performance, it renders them unsuitable for application to a different organs or modalities. Our proposed approach does not make such assumptions, is robust with respect to the exact choice of hyperparameters and could therefore effectively be transferred to new problems or datasets without requiring any modification.

We believe that our proposed method constitutes a step in improving anomaly detection for medical imaging applications. In the future anomaly detection algorithms have the potential of making use of the increasing amounts of available raw data, offering the perspective of effective radiological support tools that are not affected by the annotation data bottleneck.

\bibliographystyle{splncs04}
\bibliography{ref/refs_short2}

\begin{thebibliography}{10}
\providecommand{\url}[1]{\texttt{#1}}
\providecommand{\urlprefix}{URL }
\providecommand{\doi}[1]{https://doi.org/#1}

\bibitem{abati_and:_2018}
Abati, D., Cucchiara, R., et~al: {AND}: {Autoregressive} {Novelty} {Detectors}
  (2018)

\bibitem{alain_what_2014}
Alain, G., Bengio, Y.: What {Regularized} {Auto}-encoders {Learn} from the
  {Data}-generating {Distribution}. JMLR  (2014)

\bibitem{an_variational_2015}
An, J., Cho, S.: Variational {Autoencoder} based {Anomaly} {Detection} using
  {Reconstruction} {Probability} (2015)

\bibitem{baur_deep_2018}
Baur, C., Navab, N., et~al: Deep {Autoencoding} {Models} for {Unsupervised}
  {Anomaly} {Segmentation} in {Brain} {MR} {Images}. CoRR  (2018)

\bibitem{chen_unsupervised_2018}
Chen, X., Konukoglu, E.: Unsupervised {Detection} of {Lesions} in {Brain} {MRI}
  using constrained adversarial auto-encoders. CoRR  (2018)

\bibitem{chen_deep_2018}
Chen, X., Konukoglu, E., et~al: Deep {Generative} {Models} in the
  {Real}-{World}: {An} {Open} {Challenge} from {Medical} {Imaging}. CoRR
  (2018)

\bibitem{dai_diagnosing_2019}
Dai, B., Wipf, D.: Diagnosing and enhancing {VAE} models. In: {ICLR} (2019)

\bibitem{erihov_cross_2015}
Erihov, M., Hashoul, S., et~al: A cross saliency approach to asymmetry-based
  tumor detection. In: International {Conference} on {Medical} {Image}
  {Computing} and {Computer}-{Assisted} {Intervention}. Springer (2015)

\bibitem{goldstein_comparative_2016}
Goldstein, M., Uchida, S.: A {Comparative} {Evaluation} of {Unsupervised}
  {Anomaly} {Detection} {Algorithms} for {Multivariate} {Data}. PLoS ONE
  (2016)

\bibitem{juan-albarracin_automated_2015}
Juan-Albarracín, J., García-Gómez, J.M., et~al: Automated glioblastoma
  segmentation based on a multiparametric structured unsupervised
  classification. PLoS One  (2015)

\bibitem{kingma_auto-encoding_2013}
Kingma, D.P., Welling, M.: Auto-{Encoding} {Variational} {Bayes}. CoRR  (2013)

\bibitem{kiran_overview_2018}
Kiran, B., Parakkal, R., et~al: An {Overview} of {Deep} {Learning} {Based}
  {Methods} for {Unsupervised} and {Semi}-{Supervised} {Anomaly} {Detection} in
  {Videos}. Journal of Imaging  (2018)

\bibitem{menze_multimodal_2015}
Menze, B.H., Van~Leemput, K., et~al: The {Multimodal} {Brain} {Tumor} {Image}
  {Segmentation} {Benchmark} ({BRATS}). IEEE Trans Med Imaging  (2015)

\bibitem{nalisnick_deep_2018}
Nalisnick, E., Lakshminarayanan, B., et~al: Do {Deep} {Generative} {Models}
  {Know} {What} {They} {Don}'t {Know}? ICLR  (2019)

\bibitem{paszke_automatic_2017}
Paszke, A., Lerer, A., et~al: Automatic differentiation in {PyTorch} (2017)

\bibitem{pawlowski_unsupervised_2018}
Pawlowski, N., Glocker, B., et~al: Unsupervised {Lesion} {Detection} in {Brain}
  {CT} using {Bayesian} {Convolutional} {Autoencoders} (2018)

\bibitem{radford_unsupervised_nodate}
Radford, A., Chintala, S., et~al: Unsupervised representation learning with
  deep convolutional generative adversarial networks  (2015)

\bibitem{rezende_stochastic_2014}
Rezende, D.J., Wierstra, D., et~al: Stochastic {Backpropagation} and
  {Approximate} {Inference} in {Deep} {Generative} {Models}. In: {ICML}.
  JMLR.org (2014)

\bibitem{schlegl_unsupervised_2017}
Schlegl, T., Langs, G., et~al: Unsupervised {Anomaly} {Detection} with
  {Generative} {Adversarial} {Networks} to {Guide} {Marker} {Discovery}. In:
  {IPMI}. Springer (2017)

\bibitem{van_essen_human_2012}
Van~Essen, D.C., {WU-Minn HCP Consortium}and, et~al: The {Human} {Connectome}
  {Project}: a data acquisition perspective. Neuroimage  (2012)

\bibitem{xiao_fashion-mnist:_2017}
Xiao, H., Rasul, K., Vollgraf, R.: Fashion-{MNIST}: a {Novel} {Image} {Dataset}
  for {Benchmarking} {Machine} {Learning} {Algorithms} (2017)

\bibitem{you_unsupervised_2019}
You, S., Konukoglu, E., et~al: Unsupervised {Lesion} {Detection} via {Image}
  {Restoration} with a {Normative} {Prior}. In: International {Conference} on
  {Medical} {Imaging} with {Deep} {Learning} – {Full} {Paper} {Track} (2019)

\end{thebibliography}

\section{Supplements}

\includepdf[pages=-]{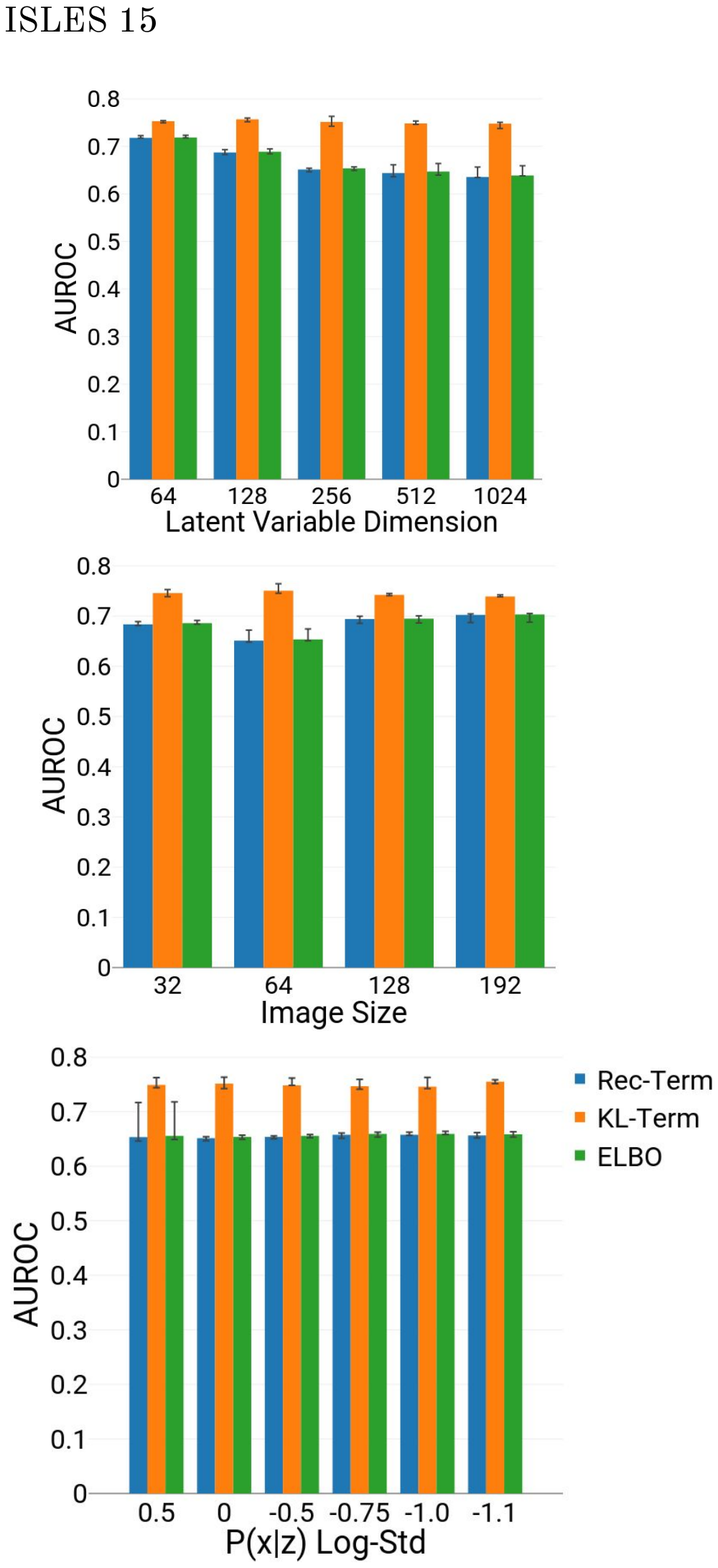}
\includepdf[pages=-]{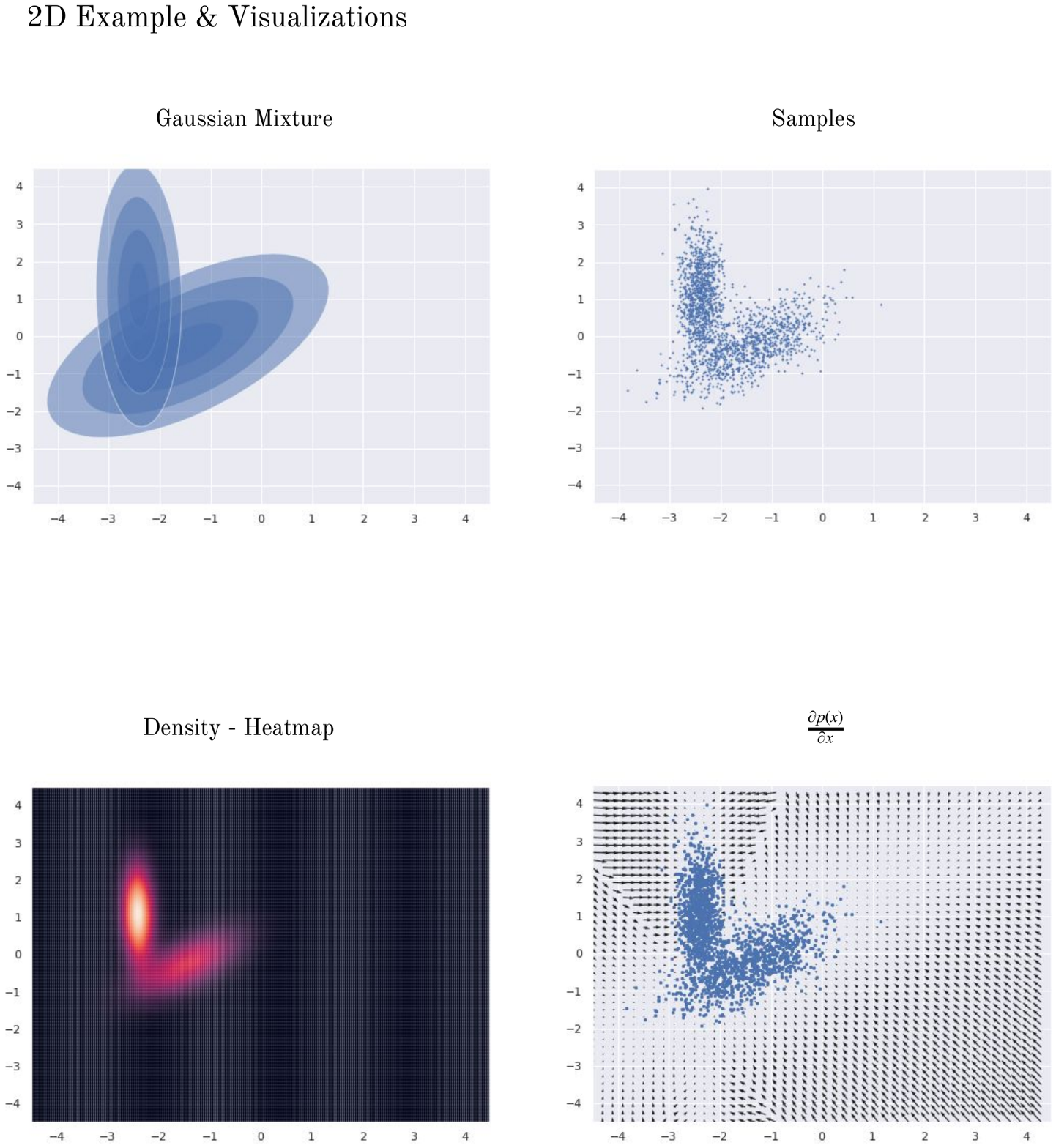}

\end{document}